\documentclass{article}
 \usepackage{graphicx} 
\usepackage{amsmath}
\usepackage{amssymb} 
\usepackage{wrapfig}
\usepackage{color}
\usepackage{soul}

\usepackage{multirow}
\usepackage{float}
\usepackage{placeins}
\usepackage{subcaption}
\usepackage{svg}
\usepackage{booktabs}
\usepackage{hyperref}
\hypersetup{
    colorlinks = true,
    urlcolor = blue,
    citecolor    = blue,
    linkcolor    = blue
    }

\usepackage[sort,numbers]{natbib}

\usepackage{xr} 
\usepackage{geometry}

\title{MMGNN: Multi-level, multi-color graph neural networks for molecular property prediction}
\author{
Trung Nguyen$^{1}$\,,\ 
and\ 
Duc Duy Nguyen$^{2}$\footnote{Address correspondence to Duc Duy Nguyen. E-mail: ducnguyen@utk.edu} \\
$^1$The Bredesen Center, University of Tennessee, Knoxville, TN 37996, USA \\
$^2$Department of Mathematics, University of Tennessee, Knoxville, TN 37996, USA
}
\begin{document}

\maketitle

\begin{abstract}
    Molecular message-passing neural networks commonly propagate chemically diverse interactions through a single graph, which may mix interaction-specific signals and require deep propagation to capture long-range effects. We introduce the Multi-level, Multi-color Graph Neural Network (MMGNN), a hierarchical framework that decomposes a molecular graph into overlapping atom-type-pair-specific subgraphs while preserving atom-level resolution. MMGNN-2D constructs chemical-colored subgraphs from covalent connectivity, whereas MMGNN-3D constructs geometric-colored subgraphs from spatial proximity and augments their edges with distance, angular, and torsional descriptors. Both variants apply a shared communicative message-passing backbone to each subgraph and combine the resulting representations through atom-wise aggregation and molecular readout. We evaluated MMGNN on five classification and three regression benchmarks from MoleculeNet using common scaffold splits and five independent runs. MMGNN-2D achieved the highest macro-average AUC-ROC of 0.838 across the classification datasets and the lowest RMSE on ESOL (0.803). MMGNN-3D obtained the highest mean AUC-ROC on BBBP (0.956) and the lowest RMSE on FreeSolv (1.793), indicating complementary strengths of topological and geometric representations. Structural and leave-one-out analyses further illustrate how the subgraph decomposition affects learned representations and atom-type-pair sensitivities. These results support overlapping interaction-specific graph decomposition as a competitive strategy for molecular property prediction.
\end{abstract}

\section{Introduction}

Accurate molecular property prediction is central to computational chemistry, drug discovery, and materials design \cite{wu2017moleculenet}. Classical quantitative structure--activity relationship models depend on predefined descriptors, whereas graph neural networks (GNNs) learn molecular representations directly from atom and bond information \cite{duvenaud2015convolutional,gilmer2017neural}. In a message-passing neural network, each atom repeatedly aggregates information from its graph neighbors, providing an effective and flexible framework for graph-level classification and regression. This paradigm has become a standard foundation for molecular representation learning.

Most molecular message-passing models nevertheless operate on a single atom--bond graph and use a shared aggregation channel for chemically diverse interactions. Their expressive power is bounded by the one-dimensional Weisfeiler--Lehman test in common formulations \cite{xu2018powerful}, and long-range communication requires increasingly deep propagation. Greater depth can blur node representations through over-smoothing \cite{li2018deeper} or compress many distant signals into fixed-size messages through over-squashing \cite{alon2020bottleneck}. Moreover, aggregation on a single graph may mix signals associated with distinct atom-type environments before their contributions can be represented separately. The resulting challenge is not merely to enlarge the receptive field, but to preserve interaction-specific information while still forming a coherent molecular representation.

Several lines of research address related aspects of this problem. Element-specific models separate interactions according to chemical element types \cite{szocinski2021awegnn}, while reduced, functional-group, motif, and hierarchical molecular graphs introduce higher-level structural representations. Recent multi-representation approaches further show that atom-, functional-group-, junction-tree-, and pharmacophore-level graphs provide complementary information \cite{kengkanna2024mmgx}. In parallel, geometric models such as SchNet and equivariant GNNs incorporate interatomic distances or coordinate-aware operations to represent three-dimensional molecular structure \cite{schutt2017schnet,satorras2021n}. These advances demonstrate the value of both chemical abstraction and molecular geometry. However, reduced-graph approaches may discard atom-level detail, and global geometric message passing can still combine chemically distinct spatial interactions within the same propagation channel.

We therefore propose the Multi-level, Multi-color Graph Neural Network (MMGNN), which represents a molecule as an overlapping collection of atom-type-pair-specific subgraphs. Unlike a disjoint fragmentation or graph-coarsening scheme, this construction retains atoms in every relevant interaction channel; an atom may participate in multiple colored subgraphs, and its subgraph-specific embeddings are subsequently combined by atom-wise aggregation. MMGNN-2D constructs chemical-colored subgraphs from covalent connectivity, whereas MMGNN-3D constructs geometric-colored subgraphs from spatial proximity and augments their edges with distance, angular, and torsional descriptors. A shared Communicative Message Passing Neural Network (CMPNN) backbone \cite{song2020communicative} processes each colored subgraph independently before the resulting representations are integrated at the atom and molecular levels. This design aims to delay the mixing of chemically distinct interactions while preserving communication through overlapping atoms and the final molecular readout.

The principal contributions of this work are threefold. First, we introduce an overlapping atom-type-pair decomposition that preserves atom-level resolution while creating interaction-specific message-passing channels. Second, we formulate complementary 2D and 3D variants within a unified architecture, enabling a controlled comparison of covalent topology and spatial geometry. Third, we evaluate MMGNN on eight MoleculeNet benchmarks using common scaffold splits and examine its behavior through aggregation ablations, a structural case study, and leave-one-out subgraph-sensitivity analyses.
Across these benchmarks, MMGNN achieves competitive classification and regression performance, with the 2D and 3D variants showing complementary strengths across molecular properties.

\section{Methodology}

\subsection{MMGNN Overview}
Standard message-passing graph neural networks (MPNNs) \cite{gilmer2017neural} operate on a single global molecular graph, where all neighboring interactions are propagated through a shared aggregation mechanism. While global molecular graphs enable information exchange across the entire molecule, all neighboring interactions are propagated through the same message-passing channel. Consequently, chemically distinct interactions may be mixed during aggregation, obscuring important local patterns.

To address this limitation, we propose the Multi-level, Multi-color Graph Neural Network (MMGNN), a hierarchical framework that decomposes a molecular graph into a collection of overlapping atom-type-specific subgraphs. Each subgraph focuses on a particular interaction type between pairs of atom categories, enabling the model to learn specialized representations for distinct chemical or geometric environments.

Importantly, MMGNN does not replace global molecular communication with isolated local fragments. The colored subgraphs form an overlapping decomposition of the original molecular graph, where the same atom may participate in multiple interaction-specific subgraphs. Through these shared atoms and the subsequent atom-wise aggregation step, information from different chemical environments can be integrated into a unified molecular representation. Thus, MMGNN preserves global molecular connectivity while reducing premature mixing of chemically distinct interactions.

Another advantage of MMGNN is its ability to improve information flow beyond immediate covalent neighborhoods. In standard 2D MPNNs, messages are propagated along covalent bonds, so information from distant atoms can only reach a target atom after multiple message-passing layers. This may lead to information attenuation, over-squashing, or over-smoothing. In contrast, the geometric-colored subgraphs in MMGNN-3D introduce spatial interaction edges between atom pairs that are close in three-dimensional space but not necessarily adjacent in the covalent graph. As a result, important non-covalent and long-range geometric interactions can be conveyed more directly. Meanwhile, MMGNN-2D preserves longer-range topological communication through overlapping atom-type-specific subgraphs and atom-wise aggregation, allowing information from different chemical environments to be integrated without prematurely mixing all interaction types in a single global channel.

\begin{figure}[h!]
    \centering
    \includegraphics[width=\textwidth]{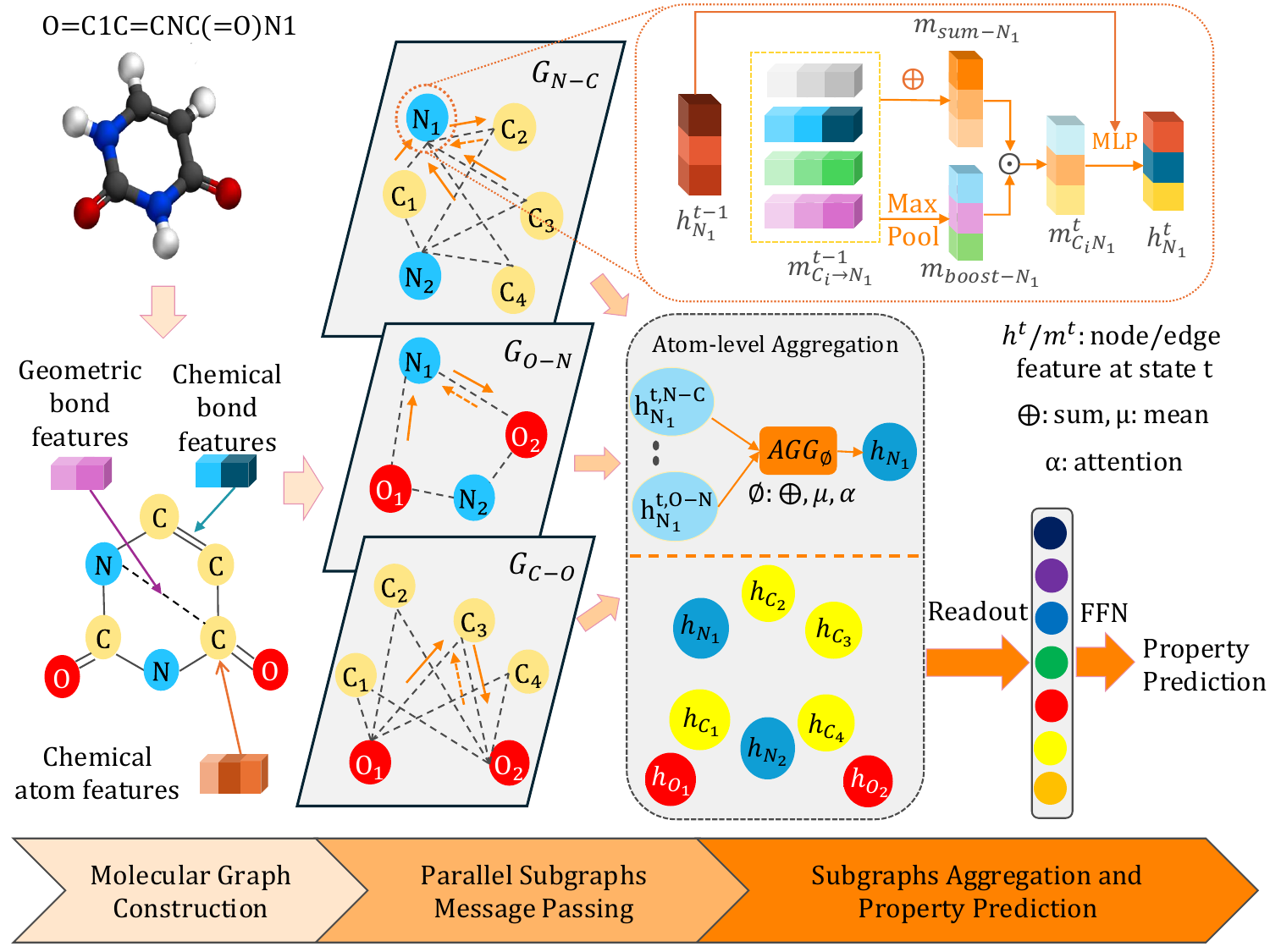}
    \caption{The MMGNN architecture. The framework proceeds in three stages: (1) Molecular graph construction: the molecule is featurized with atomic, chemical bond, and geometric bond features. (2) Parallel subgraph message passing: the molecule is decomposed into bipartite subgraphs (e.g., $G_{\text{N}-\text{C}}$, $G_{\text{O}-\text{N}}$), and each subgraph is processed independently in parallel. The top block shows how node features are updated using communicative message passing. (3) Subgraph aggregation and prediction: subgraph-specific atom embeddings are aggregated via $AGG_{\phi}$ and then read out into a global vector for property prediction.}
    \label{fig:pipeline}
\end{figure}

The overall MMGNN framework consists of three major stages, illustrated in Figure~\ref{fig:pipeline}. First, a molecule is represented by a complete atom-pair reference graph from which multiple atom-type-specific interaction subgraphs are selected. Second, each subgraph is processed independently using a shared communicative message passing neural network (CMPNN) \cite{song2020communicative} backbone to generate subgraph-specific atom embeddings. Third, atom-level embeddings from different subgraphs are aggregated to produce the final molecular representation for downstream property prediction.

To investigate the role of topology and geometry in molecular representation learning, we design MMGNN in two complementary variants:

\begin{itemize}
    \item \textbf{MMGNN-2D} constructs \emph{chemical-colored subgraphs} based on covalent bonding patterns and chemical atom-type interactions. This variant focuses on learning chemically localized topological representations without explicit 3D geometric information.
    
    \item \textbf{MMGNN-3D} constructs \emph{geometric-colored subgraphs} based on spatial proximity and geometric interactions. In addition to chemical bond features, this variant incorporates multi-body geometric descriptors including distances, bond angles, and torsion angles to capture local three-dimensional molecular environments.
\end{itemize}

By combining overlapping interaction-specific subgraphs with structured message propagation, MMGNN provides a unified framework for learning both chemically localized and globally coupled molecular representations.

\subsection{Subgraph Constructions}
\subsubsection{Molecular Graph and Atom-Type Coloring}

We represent a molecule as an atom-type-colored molecular graph 
\(\mathcal{G} = (\mathcal{V}, \mathcal{E})\), where the vertex set 
\(\mathcal{V}\) represents atoms and each atom is assigned a color according to its atom type. 
The edge set \(\mathcal{E}\) will be specified according to the type of interaction considered, including covalent chemical bonds for MMGNN-2D and spatial geometric interactions for MMGNN-3D.

For a given molecular dataset, we define a finite set of atom types
\[
\mathcal{C} = \{C_1, C_2, \ldots, C_K\},
\]
where each \(C_k \in \mathcal{C}\) denotes a distinct atom type. In datasets considered in this work, we use
\[
\mathcal{C} = \{ \text{C}, \text{H}, \text{O}, \text{N}, \text{P}, \text{Cl}, \text{F}, \text{Br}, \text{S}, \text{Si}, \text{I}, \text{X} \},
\]
where \(\text{X}\) denotes any atom type outside the first eleven categories.

Formally, the vertex set is written as
\begin{align}
    \mathcal{V} = \big\{ (\mathbf{r}_i, \alpha_i) \;\big|\; 
    \mathbf{r}_i \in \mathbb{R}^3,\;
    \alpha_i \in \mathcal{C}, \;
    i = 1,2,\ldots,N \big\},
\end{align}
where \(N\) is the number of atoms in the molecule, 
\(\mathbf{r}_i \in \mathbb{R}^3\) denotes the coordinate of atom \(i\), 
and \(\alpha_i \in \mathcal{C}\) denotes its atom type.

The complete set of possible atom pairs is given by
\begin{align}
    \mathcal{E}_{\mathrm{all}} 
    = \big\{ \{i,j\} \;\big|\; i,j \in [N],\; i \neq j \big\},
\end{align}
where \([N]=\{1,2,\ldots,N\}\). We use \(\mathcal{E}_{\mathrm{all}}\) only as a reference set from which chemically or geometrically meaningful interaction edges are selected. Specifically, MMGNN-2D restricts message passing to covalent bond edges, while MMGNN-3D selects spatial interaction edges based on geometric criteria.

This atom-type-colored representation provides the common backbone for constructing the chemical-colored and geometric-colored subgraphs described below.

\subsubsection{Chemical-colored Subgraphs}

Chemical-colored subgraphs are designed to capture covalent interaction patterns associated with specific atom-type pairs. 
While the complete set of atom pairs provides a reference for possible interactions, MMGNN-2D restricts chemical message passing to covalent bonds, which better reflects the local connectivity structure of molecular graphs.

We first define the set of covalent bond edges as
\begin{align}
    \mathcal{E}_{\mathrm{bond}} 
    = \big\{ \{i,j\} \;\big|\; \text{atoms } i \text{ and } j 
    \text{ are connected by a covalent bond in the molecule} \big\}.
\end{align}
This edge set encodes the chemical connectivity of the molecule as determined by valence structure, bond order, and aromaticity.

For a given pair of atom types \((C_k, C_{k'}) \in \mathcal{C} \times \mathcal{C}\), we define the corresponding chemical-colored subgraph as
\begin{align}
    \mathcal{G}_{kk'}^{\mathrm{chem}} 
    = \big(\mathcal{V}_{kk'}, \mathcal{E}_{kk'}^{\mathrm{chem}}\big),
\end{align}
where the vertex set \(\mathcal{V}_{kk'} \subseteq \mathcal{V}\) contains all atoms whose types belong to \(\{C_k,C_{k'}\}\):
\begin{align}
    \mathcal{V}_{kk'} 
    = \big\{ i \in [N] \;\big|\; \alpha_i \in \{C_k,C_{k'}\} \big\},
\end{align}
and the edge set is restricted to covalent bonds within this vertex subset:
\begin{align}
    \mathcal{E}_{kk'}^{\mathrm{chem}} 
    = \big\{ \{i,j\} \in \mathcal{E}_{\mathrm{bond}} 
    \;\big|\; i,j \in \mathcal{V}_{kk'} \big\}.
\end{align}

This construction yields a collection of atom-type-specific chemical subgraphs, each emphasizing covalent interactions between selected atom categories. 
Because atoms may appear in multiple chemical-colored subgraphs, the resulting decomposition preserves the connectivity of the original molecular graph while allowing MMGNN-2D to learn interaction-specific representations for different chemical environments.

\subsubsection{Geometric-colored Subgraphs}

In contrast to chemical-colored subgraphs, which are restricted to covalent bonds, geometric-colored subgraphs are designed to capture spatial interactions between atoms that may be distant in the covalent molecular graph but close in three-dimensional space. This construction allows MMGNN-3D to incorporate non-covalent and long-range geometric interactions that cannot be directly represented by standard 2D message passing along covalent bonds.

Recall that each atom \(i\) is represented by its coordinate \(\mathbf{r}_i \in \mathbb{R}^3\) and atom type \(\alpha_i \in \mathcal{C}\). For any pair of atoms \(i\) and \(j\), let
\[
    d_{ij} = \|\mathbf{r}_i-\mathbf{r}_j\|
\]
denote their Euclidean distance. We define the set of geometric interaction edges as
\begin{align}
    \mathcal{E}_{\mathrm{geom}} 
    = \big\{ \{i,j\} \;\big|\; 
    i,j \in [N],\; i \neq j,\;
    d_{\min}(i,j) \le d_{ij} \le d_{\mathrm{cut}} 
    \big\},
\end{align}
where \(d_{\mathrm{cut}}\) is a spatial cutoff radius and
\[
    d_{\min}(i,j) = r_{\mathrm{vdw},i} + r_{\mathrm{vdw},j} + \sigma
\]
is a steric lower bound determined by the van der Waals radii of atoms \(i\) and \(j\), with \(\sigma\) denoting a tolerance parameter. The lower bound removes unrealistically close atom pairs, while the upper cutoff restricts the geometric graph to spatially local but potentially non-covalent interactions.

For a given pair of atom types \((C_k,C_{k'}) \in \mathcal{C}\times\mathcal{C}\), we construct the geometric-colored subgraph
\begin{align}
    \mathcal{G}_{kk'}^{\mathrm{geom}} 
    = \big(\mathcal{V}_{kk'}, \mathcal{E}_{kk'}^{\mathrm{geom}}\big),
\end{align}
where the vertex set is defined as in the chemical case:
\begin{align}
    \mathcal{V}_{kk'} 
    = \big\{ i \in [N] \;\big|\; \alpha_i \in \{C_k,C_{k'}\} \big\},
\end{align}
and the edge set contains geometric interactions within this atom-type-specific vertex subset:
\begin{align}
    \mathcal{E}_{kk'}^{\mathrm{geom}} 
    = \big\{ \{i,j\} \in \mathcal{E}_{\mathrm{geom}} 
    \;\big|\; i,j \in \mathcal{V}_{kk'} \big\}.
\end{align}

This construction extends molecular connectivity beyond covalent bonds by directly linking atom pairs that are close in three-dimensional space. As a result, MMGNN-3D can convey spatial and non-covalent information within fewer message-passing steps, while still preserving atom-type-specific interaction channels through the colored subgraph decomposition.

\subsection{Node and Edge Features}

Each molecular graph representation is enriched with node- and edge-level descriptors that encode both chemical and geometric information. We distinguish between cheminformatics atomic and bond features, and weighted colored subgraph (WCS)-based geometric features.

\paragraph{Chemical atomic features.}
We integrate cheminformatics atomic features (CAF), denoted as $\mathcal{X}_i^{\text{CAF}}$, derived from cheminformatics tools such as RDKit \cite{landrum2006rdkit}. These include the atom type, represented by a one-hot encoding of the atomic number (100 dimensions), the number of bonds the atom is involved in (6-dimensional one-hot encoding), formal charge (5-dimensional encoding), and chirality (4-dimensional encoding capturing cases such as unspecified, tetrahedral CW/CCW, or other). Additional descriptors include the number of bonded hydrogen atoms (5-dimensional), hybridization state (sp, sp2, sp3, sp3d, or sp3d2, encoded over 5 dimensions), aromaticity (a binary indicator of whether the atom is part of an aromatic system), and atomic mass (a real-valued scalar scaled by 1/100). All categorical features are one-hot encoded to ensure consistency and numerical stability in learning \cite{heid2023chemprop}.

\paragraph{Chemical bond features.}
Along with atomic-level chemical descriptors, we define the chemical bond features, $\mathcal{X}_{ij}^{\text{Bond}}$, between atoms $i$ and $j$ to enrich the message-passing process with edge-level information \cite{heid2023chemprop}. These features, also derived from RDKit, include the bond type (single, double, triple, or aromatic, encoded in 4 dimensions), whether the bond is conjugated (1-dimensional), whether the bond is part of a ring (1-dimensional), and the bond stereochemistry (none, any, E/Z or cis/trans, encoded in 6 dimensions). All bond features are represented using one-hot encoding. 

\paragraph{Geometric edge features.}
To capture the rich spatial context of the molecular graph, we extend the edge representations beyond simple chemical features to include multi-body geometric interactions. Specifically, we encode 2-body distances, 3-body bond angles, and 4-body torsion angles into the final geometric edge feature $\mathcal{X}_{ij}^{Geom}$.

\textbf{Distance (2-body).} We encode the continuous Euclidean distance $d_{ij} = ||r_i - r_j||$ between atoms $i$ and $j$ using a Gaussian Radial Basis Function (RBF) expansion. This projects the scalar distance into a high-dimensional vector space, allowing the model to learn non-linear dependencies. We define $K$ centers $\mu_k$ uniformly spaced between 0 and a cutoff $d_{cut}$, with widths $\gamma$ determined by center spacing. The distance feature vector $e_{ij}^{RBF} \in \mathbb{R}^K$ is computed as:
$$
[e_{ij}^{RBF}]_k = \exp(-\gamma(d_{ij} - \mu_k)^2), \quad k=1,\dots,K
$$

\textbf{Angles (3-body).} To capture local angular geometry, we consider the bond angles formed by the edge $i \rightarrow k$ and the neighbors of atom $i$. For every neighbor $j \in \mathcal{N}(i) \setminus \{k\}$, we calculate the bond angle $\theta_{jik}$ via the dot product of the displacement vectors. We expand the cosine of this angle using a set of $M$ Circular Basis Functions (CBF) centered at $\mu_m$ over $[-1, 1]$:
$$
\phi_m(\theta_{jik}) = \exp\left(-\frac{(\cos(\theta_{jik}) - \mu_m)^2}{\sigma^2}\right)
$$
Since an atom may have varying numbers of neighbors, we aggregate these individual triplet vectors via summation to produce a fixed-length angular feature vector $f_{ik}^{angular} \in \mathbb{R}^M$:
$$
f_{ik}^{angular} = \sum_{j \in \mathcal{N}(i) \setminus \{k\}} [\phi_1(\theta_{jik}), \dots, \phi_M(\theta_{jik})]^\top
$$

\textbf{Torsions (4-body).} To incorporate stereochemical information and conformational geometry, we compute features based on dihedral angles. For a directed bond $j \rightarrow k$, we identify all paths of four connected atoms $i-j-k-l$. The dihedral angle $\tau_{ijkl} \in (-\pi, \pi]$ is calculated using the normal vectors of the planes defined by the bonds. Respecting the periodicity of torsion angles, we encode $\tau_{ijkl}$ using a Fourier series expansion with frequency dimension $K'$:
$$
v_{ijkl}^{torsion} = [\sin(\tau), \cos(\tau), \sin(2\tau), \cos(2\tau), \dots, \sin(K'\tau), \cos(K'\tau)]
$$
These path features are aggregated over all valid paths $\mathcal{P}_{jk}$ sharing the central bond to form the final torsion feature $f_{jk}^{torsion}$. The final geometric bond feature $\mathcal{X}_{ij}^{Geom}$ is the concatenation of these multi-body descriptors:
$$
\mathcal{X}_{ij}^{Geom} := [e_{ij}^{RBF} \parallel f_{ij}^{angular} \parallel f_{ij}^{torsion}]
$$

\paragraph{Feature composition.}
The final node and edge features depend on the type of colored subgraph. For chemical-colored subgraphs \(\{\mathcal{G}_{kk'}^{\mathrm{chem}}\}\), we use
\begin{align}
    \mathcal{X}_i &= \mathcal{X}_i^{\mathrm{CAF}}, \\
    \mathcal{X}_{ij} &= \mathcal{X}_{ij}^{\mathrm{Bond}}.
\end{align}

For geometric-colored subgraphs \(\{\mathcal{G}_{kk'}^{\mathrm{geom}}\}\), we use the same atomic descriptors and augment the edge representation with geometric features:
\begin{align}
    \mathcal{X}_i &= \mathcal{X}_i^{\mathrm{CAF}}, \\
    \mathcal{X}_{ij} &= 
    \big[
    \mathcal{X}_{ij}^{\mathrm{Bond}}
    \parallel
    \mathcal{X}_{ij}^{\mathrm{Geom}}
    \big].
\end{align}

\begin{figure}[ht]
    \centering
    \includegraphics[width=0.85\linewidth]{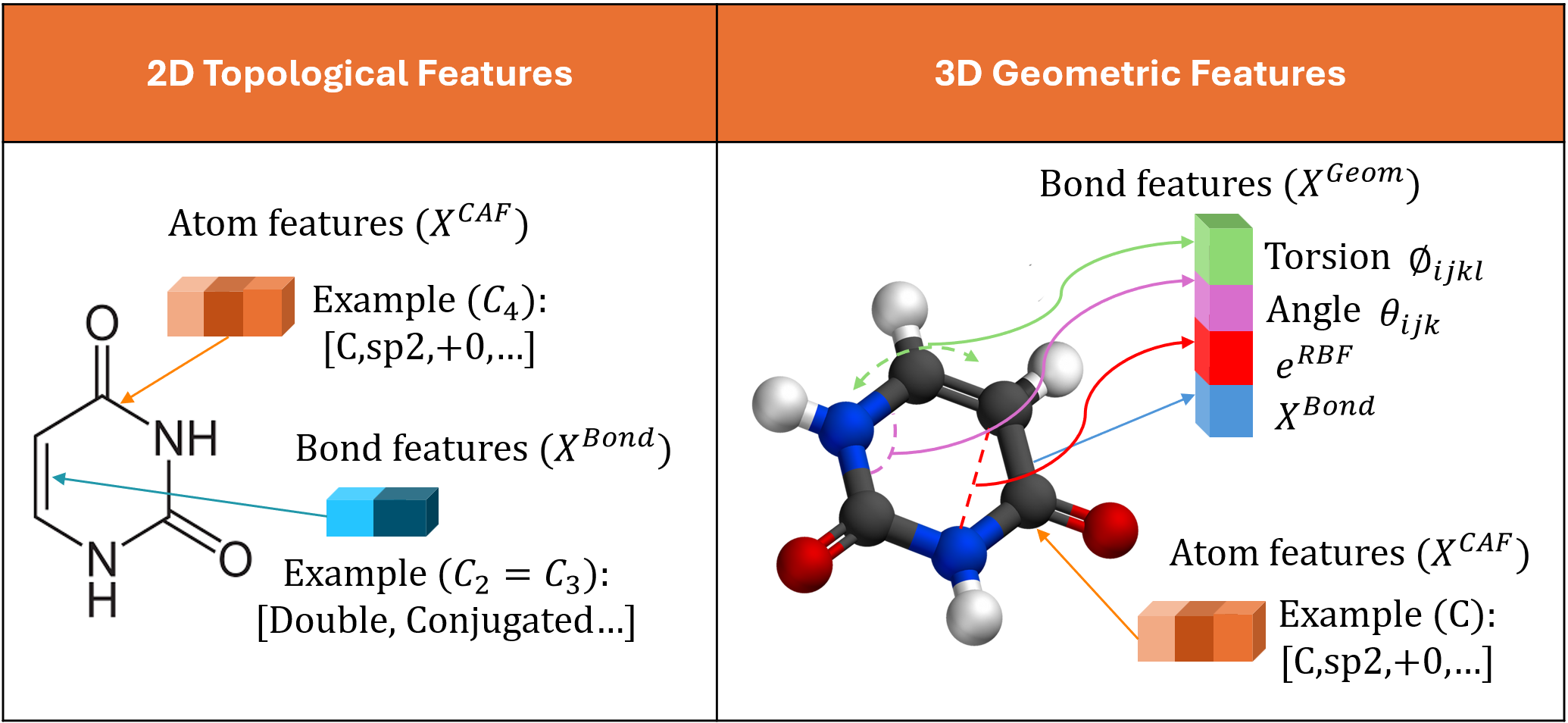}
    \caption{\textbf{Left:} 2D Topological Features including Atom features ($\mathcal{X}^{CAF}$) such as hybridization and aromaticity, and Bond features ($\mathcal{X}^{Bond}$) representing covalent connectivity. \textbf{Right:} 3D Geometric Features ($\mathcal{X}^{Geom}$) capturing spatial context via torsion angles ($\phi_{ijkl}$), bond angles ($\theta_{ijk}$), and radial basis functions ($e^{RBF}$) for distances.}
    \label{fig:features}
\end{figure}

This feature design allows MMGNN-2D to focus on covalent chemical topology, while MMGNN-3D incorporates spatial interactions, including non-covalent and topologically distant atom pairs. A visual summary of the 2D topological and 3D geometric feature sets is provided in Figure~\ref{fig:features}.

\subsection{Structured Message Passing and Prediction}

After constructing the chemical-colored and geometric-colored subgraphs, MMGNN applies a shared message-passing backbone to each subgraph independently. In this work, we use the Communicative Message Passing Neural Network (CMPNN) \cite{song2020communicative}, which updates directed edge representations using a neighbor-flow mechanism and a residual message booster. This design allows information to propagate efficiently while preserving the initial atom and edge features during iterative message passing.

Let \(\mathcal{G}_s=(\mathcal{V}_s,\mathcal{E}_s)\) denote a colored subgraph, where \(s\) indexes either a chemical-colored or geometric-colored subgraph. For a directed edge \(i\rightarrow j\) in \(\mathcal{G}_s\), let \(h_{i\rightarrow j}^{(t,s)}\) denote its hidden representation at message-passing step \(t\). The initial directed edge representation is defined by
\begin{equation}
h_{i \rightarrow j}^{(0,s)}
=
\tau \left(
W_{\mathrm{in}}
[
\mathcal{X}_{i} \parallel \mathcal{X}_{ij}
]
\right),
\end{equation}
where \(\mathcal{X}_i\) and \(\mathcal{X}_{ij}\) are the node and edge features defined in the previous section, \(\parallel\) denotes concatenation, and \(\tau\) is a nonlinear activation function.

\subsubsection*{Communicative message passing}

At each message-passing step \(t\), the incoming messages to atom \(i\) within subgraph \(\mathcal{G}_s\) are aggregated as
\begin{equation}
m_i^{(t,s)}
=
\sum_{k\in \mathcal{N}_s(i)}
h_{k\rightarrow i}^{(t-1,s)},
\end{equation}
where \(\mathcal{N}_s(i)\) denotes the neighbors of atom \(i\) in subgraph \(\mathcal{G}_s\).

For a directed edge \(i\rightarrow j\), the neighbor-flow message excludes the reverse edge \(j\rightarrow i\):
\begin{equation}
\delta_{i\rightarrow j}^{(t,s)}
=
m_i^{(t,s)}
-
h_{j\rightarrow i}^{(t-1,s)}.
\end{equation}
This update avoids immediately passing information back along the same edge and reduces redundant message loops.

The directed edge state is then updated using the CMPNN message booster:
\begin{equation}
h_{i\rightarrow j}^{(t,s)}
=
\tau\left(
W_{\mathrm{comm}}
\delta_{i\rightarrow j}^{(t,s)}
+
h_{i\rightarrow j}^{(0,s)}
\right).
\end{equation}
The residual term \(h_{i\rightarrow j}^{(0,s)}\) preserves the initial atom-edge information throughout message propagation.

After \(T\) message-passing steps, the atom representation of atom \(i\) within subgraph \(\mathcal{G}_s\) is computed as
\begin{equation}
h_{\mathrm{sub}}^{(i,s)}
=
W_{\mathrm{out}}
\left[
\mathcal{X}_i
\parallel
\sum_{j\in \mathcal{N}_s(i)}
h_{j\rightarrow i}^{(T,s)}
\right].
\end{equation}

\subsubsection*{Atom-wise subgraph aggregation}

Because the colored subgraphs form an overlapping decomposition of the molecular graph, an atom may appear in multiple subgraphs. Let \(\mathcal{S}_i\) denote the set of colored subgraphs that contain atom \(i\). MMGNN aggregates the subgraph-specific representations \(\{h_{\mathrm{sub}}^{(i,s)}:s\in\mathcal{S}_i\}\) to obtain a final atom representation \(h_i^{\mathrm{final}}\).

For sum aggregation, we use
\begin{equation}
h_i^{\mathrm{final}}
=
\sum_{s\in\mathcal{S}_i}
h_{\mathrm{sub}}^{(i,s)}.
\end{equation}
For mean aggregation, we use
\begin{equation}
h_i^{\mathrm{final}}
=
\frac{1}{|\mathcal{S}_i|}
\sum_{s\in\mathcal{S}_i}
h_{\mathrm{sub}}^{(i,s)}.
\end{equation}

For attention aggregation, the model assigns a learnable importance weight to each subgraph-specific atom representation. Specifically,
\begin{equation}
e_{i,s}
=
\mathrm{LeakyReLU}
\left(
W_{\mathrm{att}}
h_{\mathrm{sub}}^{(i,s)}
\right),
\end{equation}
and the normalized attention coefficient is
\begin{equation}
\alpha_{i,s}
=
\frac{
\exp(e_{i,s})
}{
\sum_{r\in\mathcal{S}_i}
\exp(e_{i,r})
}.
\end{equation}
The final atom representation is then given by
\begin{equation}
h_i^{\mathrm{final}}
=
\sum_{s\in\mathcal{S}_i}
\alpha_{i,s}
h_{\mathrm{sub}}^{(i,s)}.
\end{equation}

Finally, the molecular representation is obtained by summing the final atom representations:
\begin{equation}
h_{\mathrm{mol}}
=
\sum_{i\in\mathcal{V}}
h_i^{\mathrm{final}}.
\end{equation}
The molecular embedding \(h_{\mathrm{mol}}\) is passed to a feed-forward neural network for property prediction.

This message-passing and readout procedure is shared by both MMGNN variants. In MMGNN-2D, the procedure is applied to chemical-colored subgraphs constructed from covalent bond interactions and chemical bond features. In MMGNN-3D, the same procedure is applied to geometric-colored subgraphs constructed from spatial interaction edges, with edge features augmented by distance, angular, and torsional descriptors. Thus, the two variants differ only in the type of colored subgraphs and edge features used, while sharing the same subgraph-level CMPNN backbone, atom-wise aggregation, and molecular readout.

\section{Results and Discussion}
\subsection{Datasets}
We evaluated MMGNN on eight benchmark datasets from MoleculeNet \cite{wu2017moleculenet}, covering both classification and regression tasks. The classification benchmarks include BBBP, BACE, ClinTox, SIDER, and Tox21, which assess blood--brain barrier penetration, $\beta$-secretase inhibition, clinical toxicity, drug side effects, and toxicity across multiple biological targets, respectively. The regression benchmarks include Lipophilicity, FreeSolv, and ESOL, which measure octanol/water distribution coefficient, hydration free energy, and aqueous solubility, respectively.

To evaluate model generalization, we used scaffold splitting to partition each dataset into training, validation, and test sets with an 8:1:1 ratio. Scaffold splitting groups molecules according to their core chemical scaffolds, reduces scaffold overlap between partitions, and provides a more challenging assessment of generalization to structurally distinct molecules than a random split \cite{wu2017moleculenet,yang2019analyzing}. For each dataset, we performed five independent runs using random seeds from 0 to 4. To ensure a fair comparison, all baseline models were retrained using the same scaffold splits and random seeds. Final performance is reported as the mean and standard deviation across the five runs.

\subsection{Evaluation Metrics}

For classification tasks, model performance was evaluated using the area under the receiver operating characteristic curve (AUC-ROC), where higher values indicate better classification performance. For regression tasks, performance was evaluated using root mean square error (RMSE), where lower values indicate better predictive accuracy.

\subsection{Performance comparison}
To evaluate the effectiveness of the MMGNN framework, we compared our proposed MMGNN-2D and MMGNN-3D with a suite of state-of-the-art GNN architectures across eight benchmark datasets. We selected 11 baselines covering diverse GNN paradigms. GCN \cite{kipf2017semi} and GAT \cite{velivckovic2017graph} serve as standard benchmarks for graph convolution and attention mechanisms, respectively. Weave \cite{kearnes2016molecular} and SchNet \cite{schutt2017schnet} represent early architectures specifically designed for molecular representation learning. We also included powerful message-passing frameworks, including the original MPNN \cite{gilmer2017neural}, its advanced variants D-MPNN \cite{yang2019analyzing} and CMPNN \cite{song2020communicative}, which explicitly optimize edge-based message flow. Furthermore, we compared against AttentiveFP \cite{xiong2019pushing}, an attention-based fingerprint model, and recent advanced architectures such as the transformer-inspired CoMPT \cite{chen2021compt}, the dual-view CD-MVGNN \cite{ma2022cross}, and the structure-learning-based GSL-MPP \cite{yi2024molecular}. 

Unless otherwise stated, MMGNN results are reported using atom-wise sum aggregation. A detailed ablation study comparing different aggregation strategies (including mean and attention mechanisms) can be found under the section ``Comparison of Aggregation Strategies'' in the Supporting Information.

For the classification benchmarks, as presented in Table \ref{tab:classification_results}, our models achieved competitive performance across all five classification benchmarks, with MMGNN-2D obtaining the best macro-average AUC-ROC. The macro average is the unweighted mean of the dataset-level mean AUC-ROC values, giving equal importance to each dataset. MMGNN-2D demonstrated robust performance, achieving a macro-average AUC-ROC of 0.838 and the highest mean AUC-ROC on BACE ($0.877$). On ClinTox, MMGNN-2D and CD-MVGNN obtained the same rounded mean AUC-ROC ($0.913$), but MMGNN-2D had a lower standard deviation ($0.019$ versus $0.050$). MMGNN-3D obtained the highest mean AUC-ROC on BBBP ($0.956$), closely followed by MMGNN-2D ($0.955$). These results suggest that explicit 3D spatial information may be particularly beneficial for modeling blood--brain barrier penetration. Overall, MMGNN-2D achieved the strongest macro-average performance, while MMGNN-3D showed particular strength on BBBP.

\begin{table}[h]
\centering
\caption{Performance comparison (AUC-ROC $\uparrow$) on five classification datasets. Dataset-level values are reported as the mean $\pm$ standard deviation across five runs. Macro Avg. is the unweighted mean of the five dataset-level mean AUC-ROC values and is reported without a standard deviation. Best results are highlighted in \textbf{bold}, and second-best results are underlined. When displayed means are tied, the lower standard deviation is used to break the tie.}
\small
\setlength{\tabcolsep}{3pt} 
\makebox[\textwidth][c]{
\begin{tabular}{l|ccccc|c}
\hline
\textbf{Model} & \textbf{BBBP} & \textbf{BACE} & \textbf{ClinTox} & \textbf{SIDER} & \textbf{Tox21} & \textbf{Macro Avg.} \\
\hline
SchNet & $0.847 \pm 0.024$ & $0.750 \pm 0.033$ & $0.717 \pm 0.042$ & $0.545 \pm 0.038$ & $0.727 \pm 0.025$ & $0.717$ \\
Weave & $0.837 \pm 0.065$ & $0.791 \pm 0.008$ & $0.823 \pm 0.023$ & $0.543 \pm 0.034$ & $0.731 \pm 0.044$ & $0.745$ \\
GAT & $0.846 \pm 0.054$ & $0.822 \pm 0.024$ & $0.850 \pm 0.030$ & $0.528 \pm 0.002$ & $0.741 \pm 0.021$ & $0.757$ \\
GCN & $0.813 \pm 0.031$ & $0.813 \pm 0.031$ & $0.855 \pm 0.025$ & $0.598 \pm 0.014$ & $0.779 \pm 0.014$ & $0.772$ \\
MPNN & $0.896 \pm 0.041$ & $0.815 \pm 0.044$ & $0.869 \pm 0.054$ & $0.545 \pm 0.040$ & $0.768 \pm 0.024$ & $0.779$ \\
CMPNN & $0.928 \pm 0.025$ & $0.853 \pm 0.023$ & $0.898 \pm 0.022$ & $0.615 \pm 0.020$ & $0.810 \pm 0.022$ & $0.820$ \\
D-MPNN & $0.898 \pm 0.031$ & $0.852 \pm 0.053$ & $0.868 \pm 0.040$ & $0.618 \pm 0.023$ & $0.822 \pm 0.023$ & $0.812$ \\
AttentiveFP & $0.878 \pm 0.050$ & $0.793 \pm 0.015$ & $0.899 \pm 0.020$ & $0.593 \pm 0.060$ & $0.774 \pm 0.020$ & $0.787$ \\
CD-MVGNN & $0.929 \pm 0.031$ & $0.854 \pm 0.053$ & \underline{0.913 $\pm$ 0.050} & \underline{0.625 $\pm$ 0.029} & \textbf{0.828 $\pm$ 0.019} & \underline{0.830} \\
CoMPT & $0.925 \pm 0.027$ & $0.832 \pm 0.043$ & $0.878 \pm 0.029$ & $0.613 \pm 0.014$ & $0.792 \pm 0.022$ & $0.808$ \\
GSL-MPP & $0.939 \pm 0.028$ & \underline{0.858 $\pm$ 0.020} & $0.828 \pm 0.020$ & \textbf{0.648 $\pm$ 0.014} & $0.800 \pm 0.021$ & $0.815$ \\
\hline
\textbf{MMGNN-2D} & \underline{0.955 $\pm$ 0.023} & \textbf{0.877 $\pm$ 0.027} & \textbf{0.913 $\pm$ 0.019} & $0.624 \pm 0.011$ & \underline{0.823 $\pm$ 0.013} & \textbf{0.838} \\
\textbf{MMGNN-3D} & \textbf{0.956 $\pm$ 0.009} & $0.830 \pm 0.052$ & $0.902 \pm 0.036$ & $0.608 \pm 0.025$ & $0.816 \pm 0.018$ & $0.822$ \\
\hline
\end{tabular}
}
\label{tab:classification_results}
\end{table}

For the regression tasks, Table \ref{tab:regression_results} shows that the MMGNN framework continued to demonstrate strong predictive capability. MMGNN-3D achieved the lowest RMSE on FreeSolv ($1.793$), outperforming standard message-passing baselines such as CMPNN ($2.060$) and suggesting that explicit 3D geometric features may be beneficial for predicting hydration free energy. MMGNN-2D achieved the second-lowest RMSE on Lipophilicity ($0.599$) and the lowest RMSE on ESOL ($0.803$). We do not average RMSE values across these datasets because their target properties have different units and scales.

\begin{table}[h!]
\centering
\caption{Performance comparison (RMSE $\downarrow$) on three regression datasets. Values are reported as the mean $\pm$ standard deviation across five runs. RMSE values are not averaged across datasets because the target properties have different units and scales. Best results are highlighted in \textbf{bold}, and second-best results are underlined.}
\small
\setlength{\tabcolsep}{6pt} 
\begin{tabular}{l|ccc}
\hline
\textbf{Model} & \textbf{FreeSolv} & \textbf{Lipophilicity} & \textbf{ESOL} \\
\hline
SchNet & $3.215 \pm 0.755$ & $0.909 \pm 0.098$ & $1.045 \pm 0.064$ \\
Weave & $3.131 \pm 0.250$ & $0.813 \pm 0.132$ & $1.158 \pm 0.055$ \\
GAT & $2.510 \pm 0.410$ & $0.851 \pm 0.128$ & $0.942 \pm 0.062$ \\
GCN & $2.410 \pm 0.852$ & $0.763 \pm 0.080$ & $0.894 \pm 0.030$ \\
MPNN & $2.185 \pm 0.952$ & $0.712 \pm 0.051$ & $1.167 \pm 0.430$ \\
CMPNN & $2.060 \pm 0.505$ & $0.625 \pm 0.027$ & $0.838 \pm 0.140$ \\
D-MPNN & $2.271 \pm 0.508$ & $0.639 \pm 0.017$ & $0.963 \pm 0.082$ \\
AttentiveFP & $2.830 \pm 0.420$ & $0.730 \pm 0.030$ & $0.853 \pm 0.060$ \\
CD-MVGNN & $1.921 \pm 0.423$ & \textbf{0.569 $\pm$ 0.038} & $0.825 \pm 0.093$ \\
CoMPT & $2.119 \pm 0.651$ & $0.629 \pm 0.038$ & $0.829 \pm 0.075$ \\
GSL-MPP & $2.083 \pm 0.437$ & $0.694 \pm 0.070$ & $0.818 \pm 0.067$ \\
\hline
\textbf{MMGNN-2D} & \underline{1.856 $\pm$ 0.375} & \underline{0.599 $\pm$ 0.013} & \textbf{0.803 $\pm$ 0.095} \\
\textbf{MMGNN-3D} & \textbf{1.793 $\pm$ 0.367} & $0.742 \pm 0.040$ & \underline{0.814 $\pm$ 0.102} \\
\hline
\end{tabular}
\label{tab:regression_results}
\end{table}

\section{Chemical Interpretability and Subgraphs Analysis}

\subsection{Preservation of Localized Chemical Signals: A Structural Case Study}

\begin{figure}[!htbp]
\centering
\includegraphics[width=\textwidth]{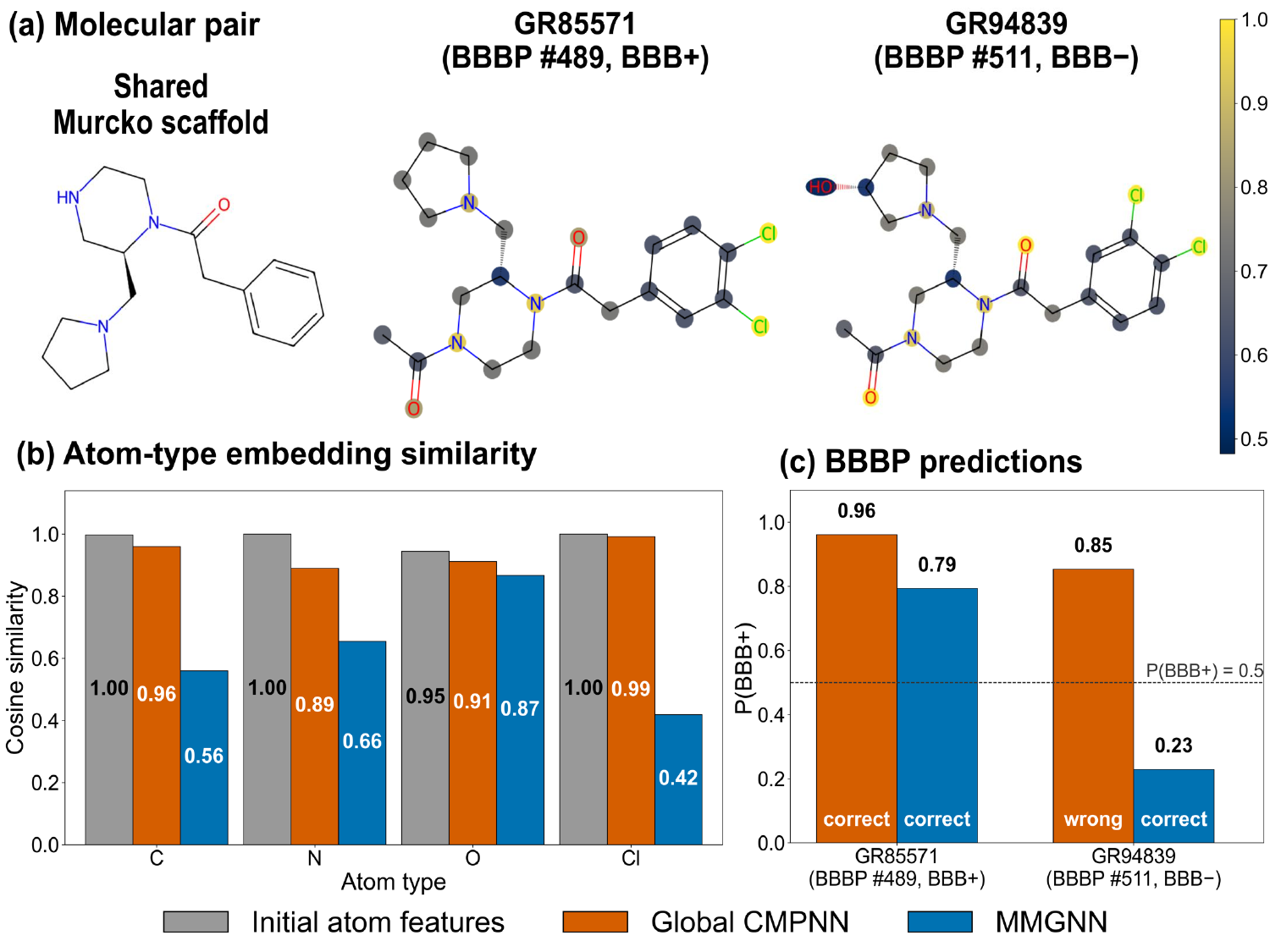}
\caption{Comparison of atom-type embedding similarities and BBBP predictions for a pair of structural analogs using a global CMPNN and MMGNN. (a) The molecular pair shares a large Murcko scaffold but possesses opposite BBBP permeability labels. (b) Cosine similarities of atom-type representations. Initial atom features exhibit similarities near 1.0 due to the shared backbone topology. Latent similarities are calculated by comparing the aggregated hidden feature vectors for each specific atom type (C, N, O, Cl) at the final network layer prior to global pooling. MMGNN produces significantly lower cosine similarities across atom types, indicating successful preservation of localized functional differences compared to the over-smoothed CMPNN representations. (c) Predictive outputs for both models, with the dashed line indicating the $P(\text{BBB}+) = 0.5$ classification threshold, MMGNN correctly classifies both analogs, whereas CMPNN incorrectly classifies the BBB- compound.}
\label{fig:similarity_prediction}
\end{figure}

A potential limitation of global message-passing networks is that signals from small but discriminative functional groups may be diluted by a large shared structural scaffold. To examine whether the subgraph decomposition in MMGNN helps preserve such localized signals, we compared its learned representations with those of a standard global CMPNN using a pair of structural analogs from the BBBP test set (Figure \ref{fig:similarity_prediction}a).

The two molecules, GR85571 (BBB+) and GR94839 (BBB-), share an identical Murcko scaffold but differ in localized functional groups (e.g., an explicit chloride and distinct nitrogen substitutions), resulting in opposite blood--brain barrier permeability labels. Because they share such a large common scaffold, their initial atom-level features, which are derived from basic topological properties such as atomic number, hybridization, and degree, are nearly identical, yielding baseline cosine similarities approaching 1.0 (Figure \ref{fig:similarity_prediction}b).

To quantify information preservation during message passing, we calculated the cosine similarity between the final-layer hidden representations of the two molecules. Specifically, the node embeddings were mean-pooled by atom type (C, N, O, Cl) at the final network layer prior to the global readout phase, and the cosine similarity was computed between the corresponding aggregated vectors.

When processed by the global CMPNN, the atom-type embeddings of the two analogs remain highly similar (e.g., Oxygen similarity $= 0.95$), suggesting that the dominant shared scaffold over-smooths the representations. In this case study, this limited separation in the feature space causes the global model to incorrectly predict the permeability of the BBB- analog (Figure \ref{fig:similarity_prediction}c).

In contrast, MMGNN decomposes the molecular graph into chemically colored subgraphs that are processed independently before atom-wise aggregation. As illustrated in Figure \ref{fig:similarity_prediction}b, this design reduces interference from the shared scaffold, producing significantly lower cosine similarities for corresponding atom types (e.g., lowering the similarity of Oxygen embeddings to 0.66). By preserving these localized chemical distinctions, the MMGNN effectively separates the analogs in the latent space and correctly predicts the probability $P(\text{BBB}+)$ for both compounds relative to the 0.5 decision threshold (Figure \ref{fig:similarity_prediction}c). Although this single example does not establish a universal solution to over-smoothing, it empirically supports the hypothesis that subgraph-based message passing better preserves localized chemical differences.

\FloatBarrier

\subsection{Subgraph Importance Analysis}

To examine whether the MMGNN framework is sensitive to distinct chemically meaningful subgraph patterns across different task domains, we expanded our interpretability analysis to the entire MoleculeNet benchmark suite. We employed a Leave-One-Out subgraph ablation strategy across both the MMGNN-2D and MMGNN-3D architectures. 

To reduce the influence of rare subgraph types, we applied a strict frequency filter: any subgraph type appearing in less than 5\% of a dataset's molecules was excluded from the ranking. For each dataset, we systematically removed a qualifying subgraph and recorded the resulting degradation in predictive performance. Focusing on the top 5 most influential subgraphs, the results were aggregated into classification tasks (BBBP, BACE, ClinTox, SIDER, Tox21) and regression tasks (FreeSolv, Lipophilicity, ESOL) to summarize the atom-type-pair sensitivities learned by the models.

\begin{figure}[h!]
    \centering
    \begin{subfigure}[b]{0.49\textwidth}
        \centering
        \includegraphics[width=\textwidth]{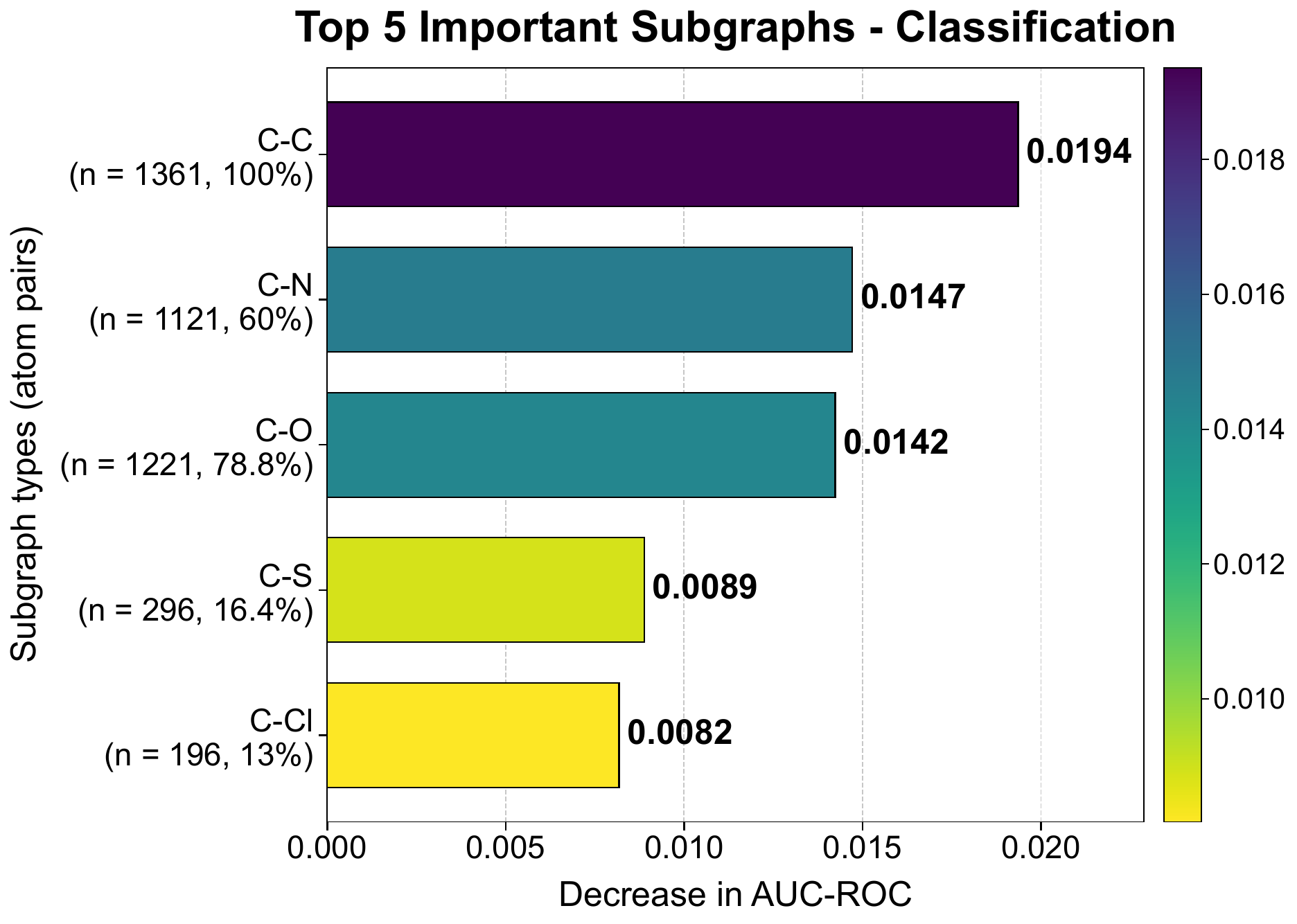}
        \caption{MMGNN-2D: Classification}
        \label{fig:loo_2d_class}
    \end{subfigure}
    \hfill
    \begin{subfigure}[b]{0.49\textwidth}
        \centering
        \includegraphics[width=\textwidth]{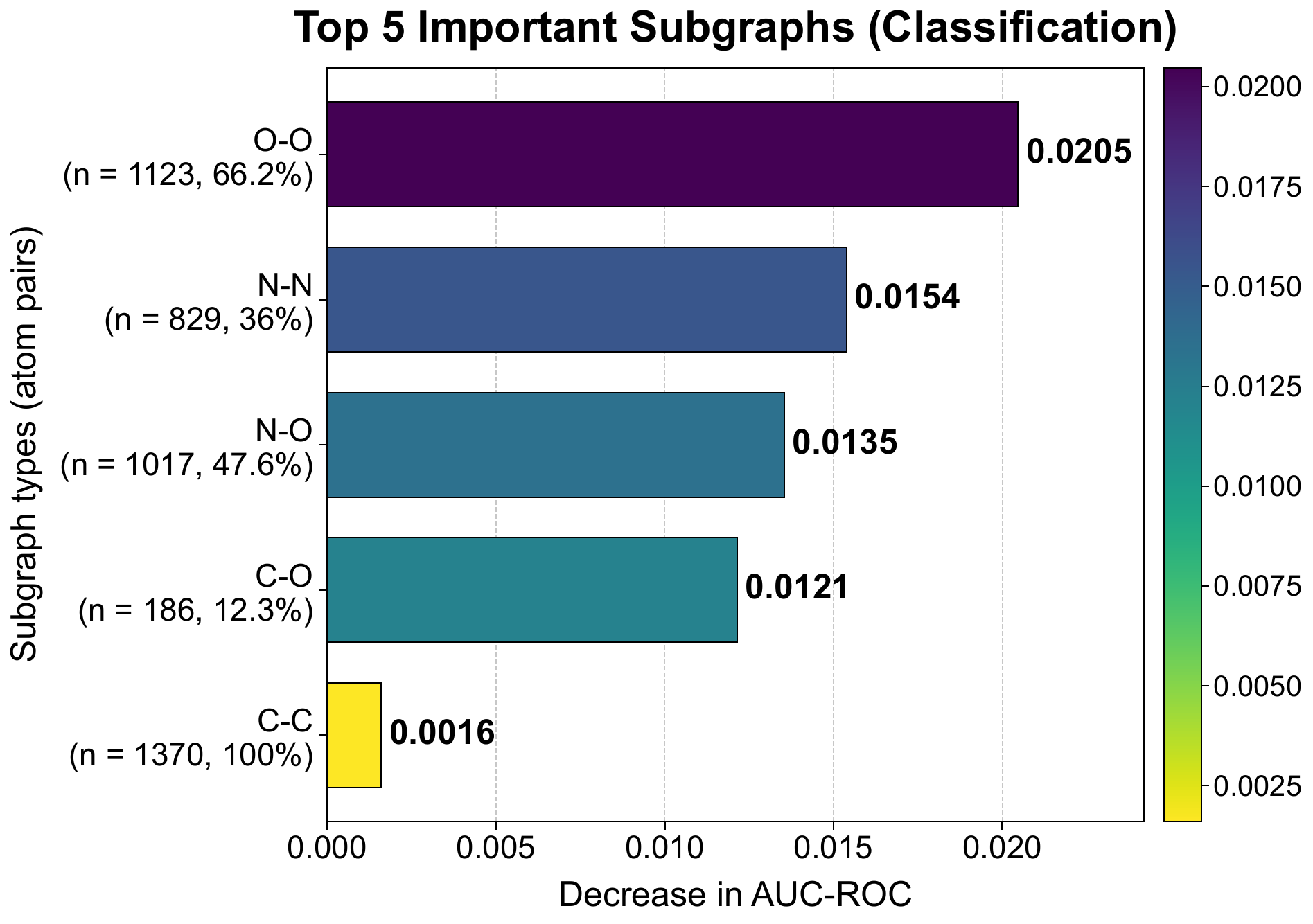}
        \caption{MMGNN-3D: Classification}
        \label{fig:loo_3d_class}
    \end{subfigure}
    \caption{Top 5 Subgraph Importance for Classification Tasks. The 2D architecture shows high sensitivity to core structural framework pairs (C-C, C-N) and basic connectivity, whereas the 3D architecture shows increased sensitivity to spatial heteroatom pairs (O-O, N-N).}
    \label{fig:loo_classification}
\end{figure}

For the classification tasks, as illustrated in Figure \ref{fig:loo_classification}, the topological MMGNN-2D model shows high sensitivity to the fundamental organic framework and standard functional connectivity. Its top subgraphs (C-C, C-N, C-O, C-S, C-Cl) are consistent with contributions from molecular scaffold structure and common polar functional environments. Because the 2D model does not use explicit spatial geometry, these sensitivities likely reflect the information available from covalent topology.

In contrast, the spatially-aware MMGNN-3D model shows greater sensitivity to dense, electron-rich heteroatom spatial pairs. Its top subgraphs are dominated by O-O, N-N, and N-O, alongside the common C-O and C-C pairs. For classification tasks involving biological activity or permeability, this pattern suggests that the model uses spatial arrangements involving polar or electron-rich regions as informative features, rather than relying only on the carbon scaffold.

When evaluating continuous thermodynamic properties (e.g., hydration free energy and aqueous solubility), the baseline error scales vary across FreeSolv, Lipophilicity, and ESOL. To account for this, we standardized the predictive impact metric to measure the percentage increase in Root Mean Square Error (\% $\Delta$ RMSE) upon subgraph ablation.

\begin{figure}[h!]
    \centering
    \begin{subfigure}[b]{0.49\textwidth}
        \centering
        \includegraphics[width=\textwidth]{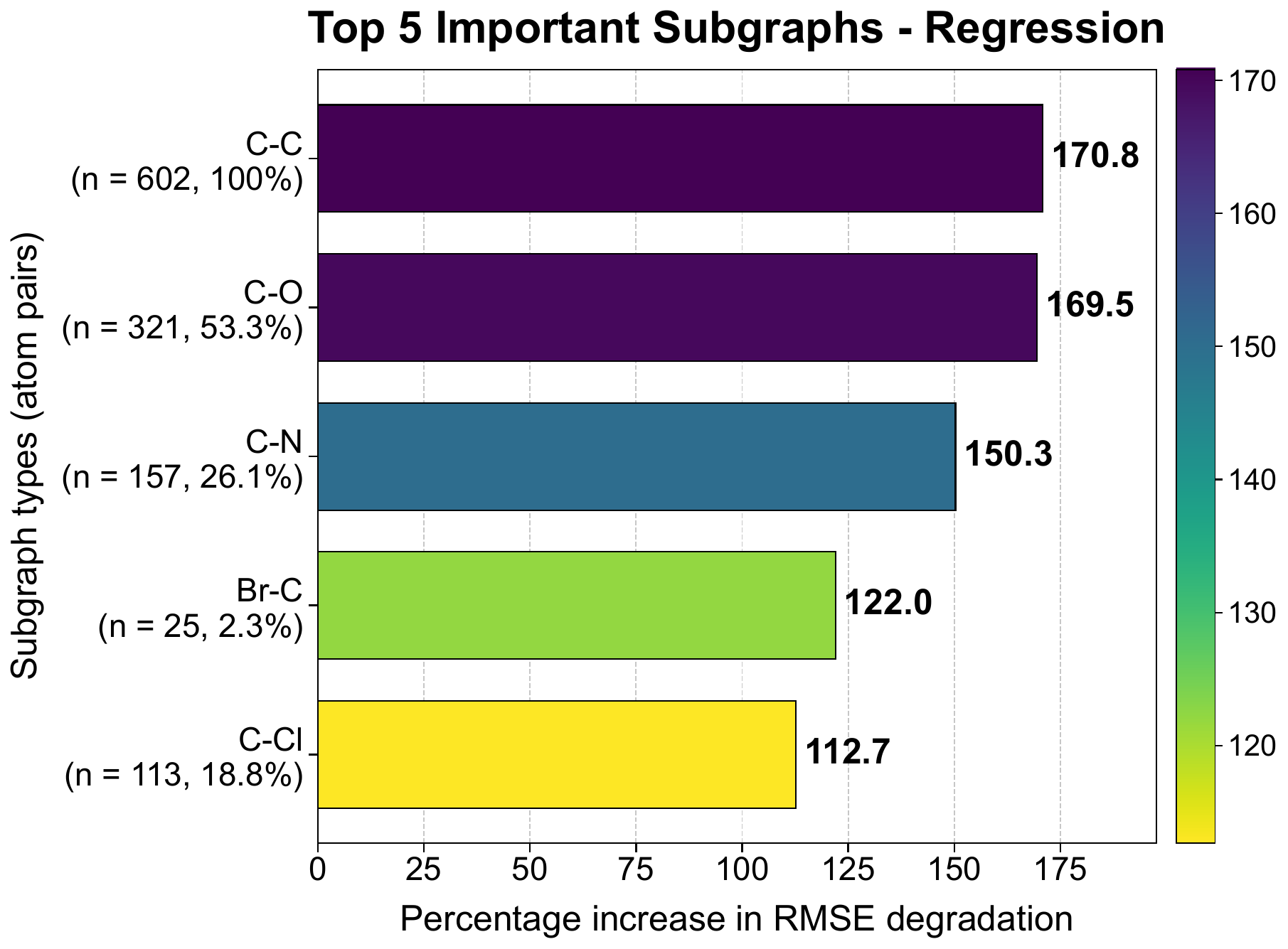}
        \caption{MMGNN-2D: Regression}
        \label{fig:loo_2d_reg}
    \end{subfigure}
    \hfill
    \begin{subfigure}[b]{0.49\textwidth}
        \centering
        \includegraphics[width=\textwidth]{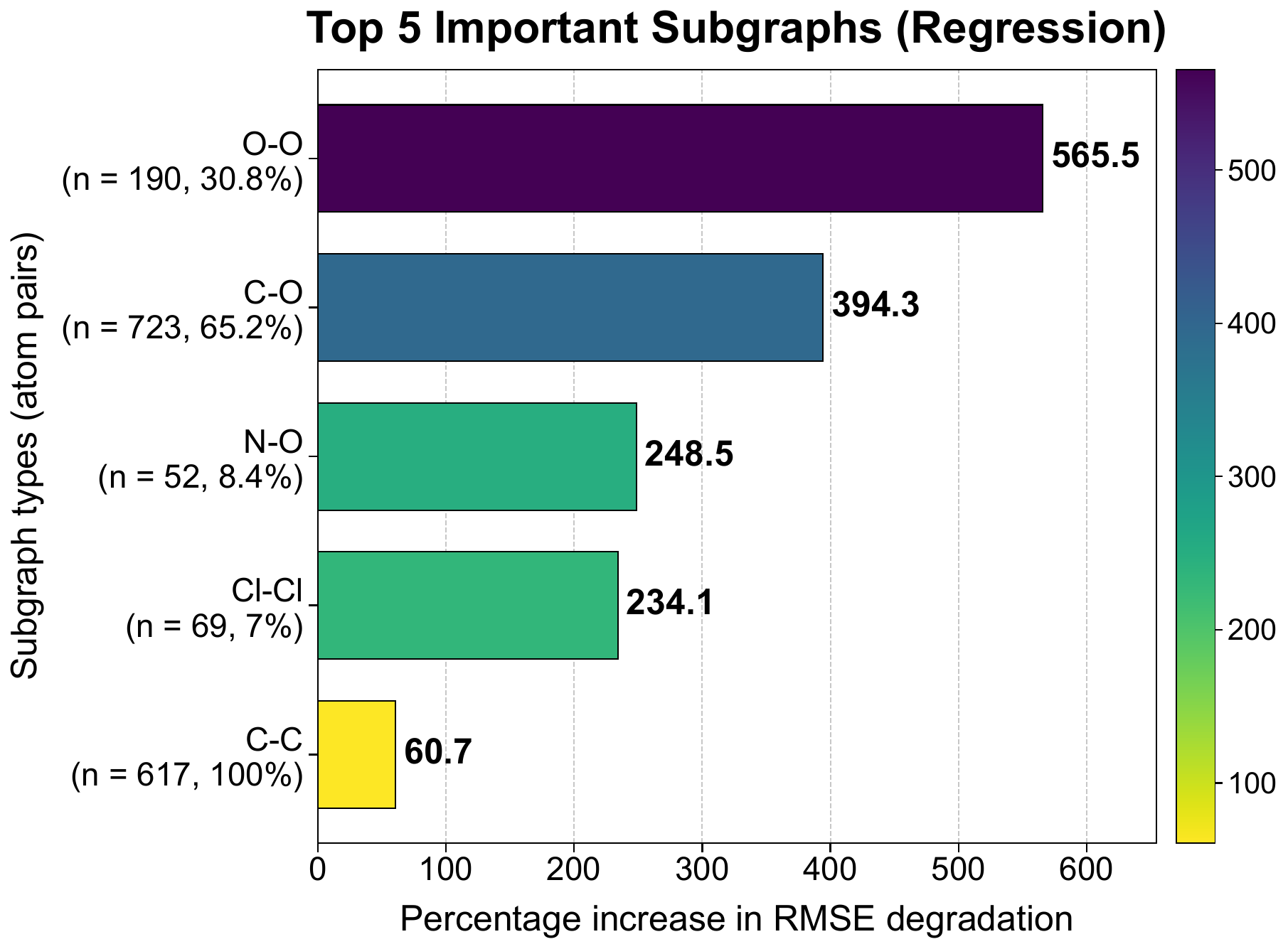}
        \caption{MMGNN-3D: Regression}
        \label{fig:loo_3d_reg}
    \end{subfigure}
    \caption{Top 5 Subgraph Importance for Regression Tasks. Evaluated by normalizing the percentage increase in RMSE. The 2D architecture shows sensitivity to polar motifs (C-O, C-N) and halogenated pairs (Br-C, C-Cl), whereas the 3D architecture shows sensitivity to spatial heteroatom pairings (O-O, N-O).}
    \label{fig:loo_regression}
\end{figure}

As shown in Figure \ref{fig:loo_regression}, the MMGNN-2D regression profile is most sensitive to C-C, C-O, C-N, Br-C, and C-Cl. This hierarchy is chemically plausible because carbon scaffolds, polar heteroatom environments, and halogenated motifs are all relevant to solvation and lipophilicity, although the ablation ranking should be interpreted as model sensitivity rather than direct causal attribution.

Mirroring the classification results, the MMGNN-3D regression profile assigns high importance to polar spatial pairings. The prominence of O-O, C-O, N-O, Cl-Cl, and C-C is consistent with the relevance of three-dimensional heteroatom environments, electronegative regions, and molecular scaffold structure to solvation-related properties. The recurrence of C-C and C-O across the top-5 profiles suggests that these atom-type pairs are broadly informative in this benchmark suite, while the remaining subgraph sensitivities vary with the property and with the availability of 3D spatial information.

For a comprehensive dataset-by-dataset analysis of all eight MoleculeNet benchmarks, see Figures S3--S10 in the Supporting Information.

\section{Conclusion}
In this work, we introduced the Multi-level, Multi-color Graph Neural Network (MMGNN), a hierarchical framework for molecular property prediction that decomposes molecular graphs into overlapping atom-type-specific subgraphs. Both MMGNN variants use a CMPNN backbone to learn subgraph-level representations and aggregate them into atom- and molecule-level embeddings. MMGNN-2D operates on covalent connectivity, whereas MMGNN-3D incorporates spatial interactions through distance, angular, and torsional descriptors.

Across eight MoleculeNet benchmarks evaluated using common scaffold splits, MMGNN-2D achieved the highest macro-average AUC-ROC of 0.838 over the five classification datasets. It also obtained the lowest RMSE on ESOL and the second-lowest RMSE on Lipophilicity and FreeSolv. MMGNN-3D achieved the highest mean AUC-ROC on BBBP and the lowest RMSE on FreeSolv, suggesting that explicit geometric information may be particularly beneficial for selected molecular properties. Because the regression targets have different units and scales, we did not average RMSE values across datasets.

The structural case study and leave-one-out analyses further illustrate how subgraph decomposition affects the representations and atom-type-pair sensitivities learned by the models. These analyses should be interpreted as evidence of model behavior rather than causal identification of universal chemical drivers. Overall, the results support atom-type-specific graph decomposition as a competitive approach to molecular representation learning. Future work will evaluate the framework on larger molecular systems, including protein--ligand complexes, and investigate self-supervised pretraining with unlabeled chemical data.

\section*{Associated Content}
\textbf{Supporting Information.}
Additional aggregation-strategy comparisons, geometric-feature ablation studies, and dataset-specific leave-one-out subgraph importance analyses are available in the Supporting Information.

\section*{Author Information}
\textbf{Corresponding Author}

Duc Duy Nguyen -- Department of Mathematics, University of Tennessee, Knoxville, TN 37996, USA; Email: \href{mailto:ducnguyen@utk.edu}{ducnguyen@utk.edu}

\textbf{Author Contributions}

T.N. and D.D.N. designed the study. T.N. implemented the method, performed experiments, analyzed results, and drafted the manuscript. D.D.N. supervised the project and revised the manuscript. Both authors reviewed and approved the final manuscript.

\textbf{Notes}

The authors declare no competing financial interest.

\section*{Data and Software Availability}
The source code is available at the GitHub repository: \url{https://github.com/MathIntelligence/MMGNN}. The benchmark datasets used in this study are publicly available through MoleculeNet. Details of dataset splits, model settings, and supporting analyses are provided in the manuscript and Supporting Information.

\section*{Acknowledgments}
This work is supported in part by funds from the National Science Foundation (NSF: \#2516126, \#2151802, and \#2534947).





\bibliographystyle{unsrtnat}
\bibliography{refs}

\end{document}